\documentclass[10pt,twocolumn,letterpaper]{article}

\usepackage{stylee}
\usepackage{times}
\usepackage{epsfig}
\usepackage{graphicx}
\usepackage{amsmath}
\usepackage{amssymb}
\usepackage{subcaption}
\usepackage[lined,ruled,commentsnumbered]{algorithm2e}
\usepackage{makecell}
\usepackage{multirow}
\usepackage{arydshln}
\usepackage{url}
\usepackage{float}
\usepackage{enumitem}
\usepackage{footnote}

\usepackage{refcount}

\newcommand{\myfootnotetext}[1]{\footnotetext{#1\label{fn:text}%
        \edef\fnmark{\getpagerefnumber{fn:mark}}%
        \edef\fntext{\getpagerefnumber{fn:text}}%
        \ifx\fnmark\fntext\else\ClassWarning{}{footnote mark and text on different pages!}\fi}}

\makeatletter
\newcommand{\removelatexerror}{\let\@latex@error\@gobble}
\makeatother

\usepackage{color}

\newcolumntype{K}{>{\centering\arraybackslash}m{2.5cm}}
\newcolumntype{M}{>{\centering\arraybackslash}m{0.96cm}}

\graphicspath{{./images/}} 

\usepackage[pagebackref=true,breaklinks=true,letterpaper=true,colorlinks,bookmarks=false]{hyperref}

\styleefinalcopy

\ifstyleefinal\fi

\begin{document}

\title{Excitation Backprop for RNNs}

\author{Sarah Adel Bargal$^*$\textsuperscript{1}, Andrea Zunino\thanks{Equal contribution} \textsuperscript{$\;$2}, Donghyun Kim\textsuperscript{1}, Jianming Zhang\textsuperscript{3},\\ Vittorio Murino\textsuperscript{2,4}, Stan Sclaroff\textsuperscript{1}\\~\\ 
\textsuperscript{1}Department of Computer Science, Boston University  
\textsuperscript{2}Pattern Analysis \& Computer Vision (PAVIS),\\ Istituto Italiano di Tecnologia 
\textsuperscript{3}Adobe Research
\textsuperscript{4}Computer Science Department, Universit\`a di Verona\\
{\tt\small \{sbargal,donhk,sclaroff\}@bu.edu, \{andrea.zunino,vittorio.murino\}@iit.it, jianmzha@adobe.com}}

\makeatletter
\let\@oldmaketitle\@maketitle
\renewcommand{\@maketitle}{\@oldmaketitle
  \includegraphics[width=1\linewidth,height=0.23\linewidth,trim={0.3cm 0.05cm 0.4cm 0.05cm},clip]{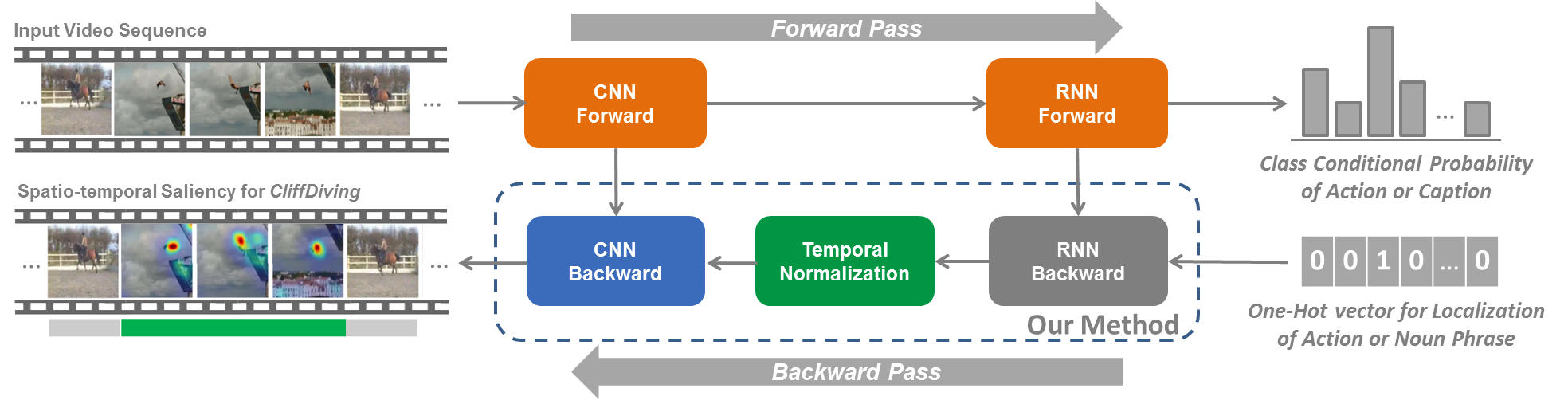}
  \vspace*{-1.3em}
  \captionof{figure}{\small{Our proposed framework spatiotemporally highlights/grounds the evidence that an RNN model used in producing a class label or caption for a given input video. In this example, by using our proposed back-propagation method, the evidence for the activity class {\em CliffDiving} is highlighted in a video that contains {\em CliffDiving} and {\em HorseRiding}. Our model employs 
  a single backward pass to produce saliency maps that highlight the evidence that a given RNN used in generating its outputs. }}
  \vspace*{-1em}
  \label{fig:pipeline}
  \bigskip\bigskip}
\makeatother

\maketitle



\begin{abstract}
\vspace*{-0.2em}
Deep models are state-of-the-art for many vision tasks including video action recognition and video captioning. Models are trained to caption or classify activity in videos, but little is known about the evidence used to make such decisions. Grounding decisions made by deep networks has been studied in spatial visual content, giving more insight into model predictions for images. However, such studies are relatively lacking for models of spatiotemporal visual content \textendash \; videos.
In this work, we devise a formulation that simultaneously grounds evidence in space and time, in a single pass, using top-down saliency. We visualize the spatiotemporal cues that contribute to a deep model's classification/captioning output using the model's internal representation. Based on these spatiotemporal cues, we are able to localize segments within a video that correspond with a specific action, or phrase from a caption, without explicitly optimizing/training for these tasks. 

\end{abstract}



\vspace*{-1em}
\section{Introduction}


To visualize what in a video gives rise to an output of a deep recurrent network, it is important to consider space and time saliency, \ie, where and when. The visualization of what a deep recurrent network finds salient in an input video can enable interpretation of the model's behavior in action classification, video captioning, and other tasks. Moreover, estimates of the model's attention (\eg, saliency maps) can be used directly in localizing a given action within a video or in localizing the portions of a video that correspond to a particular concept within a caption.

Several works address visualization of model attention in Convolutional Neural Networks (CNNs) for image classification~\cite{cao2015look,zhang2016top,grad,deconv,simonyan2013deep,CAM,gradcam}. These methods produce saliency maps that visualize the importance of class-specific image regions (spatial localization). Analogous methods for Recurrent Neural Network (RNN)-based models must handle more complex recurrent, non-linear, spatiotemporal 
dependencies; thus, progress on RNNs has been limited to~\cite{karpathy2015visualizing,Ramanishka2017cvpr}. Karpathy \etal~\cite{karpathy2015visualizing} visualize the role of 
Long Short Term Memory (LSTM) cells for text input, but not for visual data. Ramanishka \etal \cite{Ramanishka2017cvpr} map words to regions in the video captioning task by dropping out (exhaustively or by sampling) video frames and/or parts of video frames to obtain saliency maps.  This can be computationally expensive, and does not consider temporal evolution but only frame-level saliency.


In contrast, 
we propose the first one-pass formulation for visualizing spatiotemporal attention in RNNs, without selectively dropping or sampling frames or frame regions. In our proposed approach, \textit{contrastive} Excitation Backprop for RNNs ($c$EB-R), we address how to ground\footnote{In this work we use the terms \textit{ground} and \textit{localize} interchangeably.} decisions of deep recurrent networks in space and time simultaneously, using top-down saliency. Our approach models the top-down attention mechanism of deep models to produce interpretable and useful task-relevant saliency maps. Our saliency maps are obtained implicitly without the need to re-train models, unlike models that include explicit attention layers~\cite{xu2015show,yao2015describing}. 
Our method does not require a model trained using explicit spatial (region/bounding box) or temporal (frame) supervision. 

Fig.~\ref{fig:pipeline} gives an overview of our approach that produces saliency maps which enable us to visualize where and when an action/caption is occurring in a video. Given a trained model, we perform the standard forward pass. In the backward pass, we use $c$EB-R to compute and propagate winning neuron probabilities normalized over space and time.  This process yields spatiotemporal attention maps. 
Our demo code is publicly available \footnote{\scriptsize{\url{https://github.com/sbargal/Caffe-ExcitationBP-RNNs}}}.

We evaluate our approach on two models from the literature: a CNN-LSTM trained for video action recognition, and a CNN-LSTM-LSTM (encoder-decoder) trained for video captioning. 
In addition, we show how the spatiotemporal saliency maps produced for these two models can be utilized for localization of segments within a video that correspond to specified activity classes or noun phrases. 




In summary, our contributions are: 
\begin{itemize}[leftmargin=*]
\itemsep0em 
\item
We are the first to formulate top-down saliency in deep recurrent models for space-time grounding of videos. 
\item
We do so using a \textit{single} \textit{contrastive} Excitation Backprop pass of an already trained model. 
\item
Although we are not directly optimizing for localization (no training is performed on spatial or temporal annotations), we show that the internal representation of the model can be utilized to perform localization.
\end{itemize}

\section{Related Work}

Several works in the literature give more insight into CNN model predictions, \ie, the \textit{evidence} behind deep model predictions. 
Such approaches are mainly devised for image understanding and can identify the importance of class-specific image regions by means of saliency maps in a weakly-supervised way. 

\textbf{Spatial Grounding.} Ribeiro \etal \cite{ribeiro2016should} explained classification predictions with applications on images. Fong \etal \cite{8237633} addressed spatial grounding in images by exhaustively perturbing image regions. Guided Backpropagation \cite{grad} and Deconvolution \cite{deconv,simonyan2013deep} used different variants of the standard backpropagation error and visualized salient parts at the image pixel level. In particular, starting from a high-level feature map, \cite{deconv} inverted the data flow inside a CNN, from neuron activations in higher layers down to the image level. Guided Backpropagation \cite{grad} introduced an additional guidance signal to standard backpropagation preventing backward flow of negative gradients. Simonyan \etal \cite{simonyan2013deep} directly computed  the gradient of the class score with respect to the image pixel to find the spatial cues that help the class predictions in a CNN. CAM \cite{CAM} removed the last fully connected layer of a CNN and exploited a weighted sum of the last convolutional feature maps to obtain the class activation maps. 
Zhang \etal \cite{zhang2016top} generated class activation maps from any CNN architecture that uses non-linearities producing non-negative activations. Oquab \etal \cite{Oquab_2015_CVPR} used mid-level CNN outputs on overlapping patches, requiring multiple passes through the network. 

\textbf{Spatiotemporal Grounding.} Weakly-supervised visual saliency is much less explored for temporal architectures. Karpathy \etal \cite{karpathy2015visualizing} visualized interpretable LSTM cells that keep track of long-range dependencies such as line lengths, quotes, and brackets in a character-based model. Li \etal \cite{li2016visualizing} visualized a unit's salience for NLP. Selvaraju \etal \cite{gradcam} qualitatively present grounding for captioning and visual question answering in images using an RNN. Ramanishka \etal \cite{Ramanishka2017cvpr} explored visual saliency guided by captions in an encoder-decoder model. In contrast, our approach models the top-down attention mechanism of CNN-RNN models to produce interpretable and useful task-relevant spatiotemporal saliency maps that can be used for action/caption localization in videos.


\section{Background: Excitation Backprop}
In this section, a brief background on Excitation Backprop (EB) \cite{zhang2016top} is given. EB was proposed for CNNs in that work. In general, the forward activation of neuron $a_j$ in a CNN is computed by $\widehat{a}_j=\phi(\sum_iw_{ij}\widehat{a}_i+b_i)$, where $\widehat{a}_i$ is the activation coming from a lower layer, $\phi$ is a nonlinear activation function, $w_{ij}$ is the weight from neuron $i$ to neuron $j$, and $b_i$ is the added bias at layer $i$. The EB framework makes two key assumptions about the activation $\widehat{a}_j$ which are satisfied in the majority of modern CNNs due to wide usage of the \textit{ReLU} non-linearity: \textbf{A1.} $\widehat{a}_j$ is non-negative, and \textbf{A2.} $\widehat{a}_j$ is a response that is positively correlated with its confidence of the detection of specific visual features.

\begin{figure}[t]
\centering
\includegraphics[width=\linewidth, trim={0.2cm 2.8cm 0cm 0cm},clip]{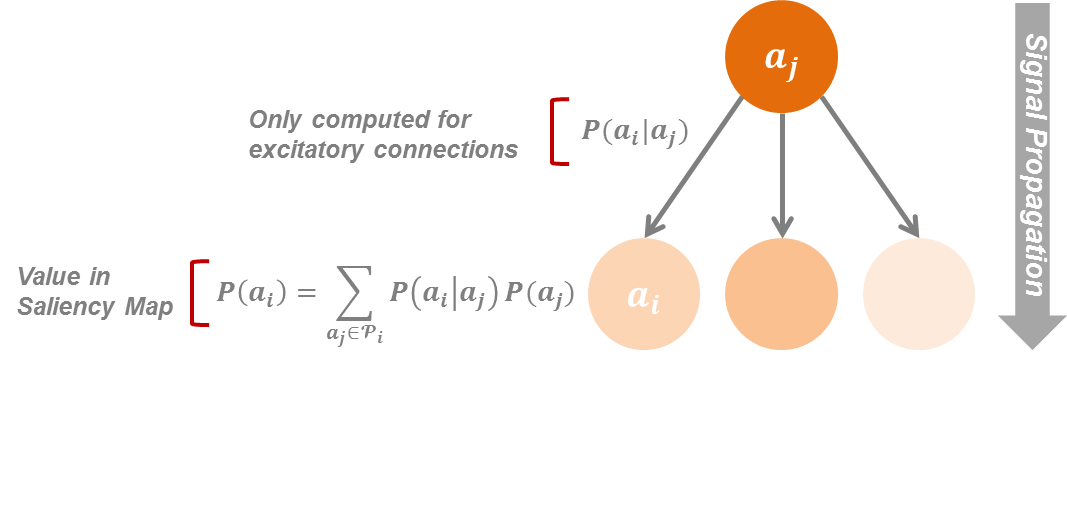}
\caption{
\small{In Excitation Backprop, excitation probabilities are propagated in a single backward pass in the CNN. A top-down signal is a probability distribution over the output units. The probabilities are backpropagated from every parent node to its children through its excitatory connections. The figure illustrates the contributions of a single parent neuron to the excitation probabilities computed at the next layer. Each $P(a_i)$  in the saliency map is computed over the complete parent set $\mathcal{P}_i$. Shading of nodes in the figure conveys $P(a_i)$ (darker shade = greater $P(a_i)$).}
\vspace{-1em}
}
\label{fig:WTA}
\end{figure}

EB realized a probabilistic Winner-Take-All formulation to efficiently compute the probability of each neuron recursively using conditional winning probabilities $P(a_i|a_j)$, normalized $\widehat{a}_i w_{ij}$ (Fig.~\ref{fig:WTA}). The top-down signal is a prior distribution over the output units. 
EB passes top-down signals through excitatory connections having non-negative weights, excluding from the competition inhibitory ones. Recursively propagating the top-down signal and preserving the sum of backpropagated probabilities layer by layer, it is possible to compute task-specific saliency maps from any intermediate layer in a single backward pass.

To improve the discriminativeness of the saliency maps, \cite{zhang2016top} introduced \textit{contrastive} EB ($c$EB) which cancels out common winner neurons and amplifies the class discriminative neurons. To do this, given an output unit $o_i \in O$, a dual unit $\overline{o}_i \in \overline{O}$ is virtually generated, whose input weights are the negation of those of $o_i$. By subtracting the saliency map for $\overline{o}_i$ from the one for $o_i$ the result better highlights cues in the image that are unique to the desired class. 

\vspace{-1mm}
\section{Our Framework}
\label{$c$EB-R}
\vspace{-1mm}

In this section we explain the details of our spatiotemporal grounding framework: $c$EB-R.  As illustrated in Fig.~\ref{fig:pipeline}, we have three main modules: RNN Backward, Temporal normalization, and CNN Backward.

\textbf{RNN Backward.} 
This module implements an excitation backprop formulation for RNNs. Recurrent models such as LSTMs are well-suited for top-down temporal saliency as they explicitly propagate information over time. 
The extension of EB for Recurrent Networks, EB-R, is not straightforward since EB must be implemented through the unrolled time steps of the RNN and since the original RNN formulation contains $tanh$ non-linearities which do not satisfy the EB assumptions \textbf{A1} and \textbf{A2}. \cite{greff2016lstm,jozefowicz2015empirical} have conducted an analysis over variations of the standard RNN formulation, and discovered that different non-linearities performed similarly for a variety of tasks. This is also reflected in our experiments. Based on this, we use \textit{ReLU} nonlinearities and corresponding derivatives, instead of $tanh$.
This satisfies \textbf{A1} and \textbf{A2}, and gives similar performance on both tasks. 

Working backwards from the RNN's output layer, we compute the conditional winning probabilities from the set of output nodes $O$, and the set of dual output nodes $\overline{O}$:
\vspace{-2mm}
\begin{equation}
P^t(a_i|a_j) = 
\begin{cases}
    Z_j \widehat{a}_i^t w_{ij},     & \text{if } w_{ij}\geq 0,     \\
    0,                              & \text{otherwise}.
\end{cases}
\label{eqn: lstm_plus}
\end{equation}
\begin{equation}
\overline{P}^t(a_i|a_j) = 
\begin{cases}
    Z_j \widehat{a}_i^t \overline{w}_{ij},     & \text{if } \overline{w}_{ij}\geq 0,     \\
    0,                              & \text{otherwise}.
\end{cases}
\label{eqn: lstm_minus}
\end{equation}
\noindent $Z_j = 1/\sum_{i:w_{ij \ge 0}}{\hat{a}_i^t w_{ij}}$ is a normalization factor such that the sum of all conditional probabilities of the children of $a_j$ (Eqn.s \ref{eqn: lstm_plus}, \ref{eqn: lstm_minus}) sum to 1; $w_{ij} \in W$ where $W$ is the set of model weights and $w_{ij}$ is the weight between child neuron $a_i$ and parent neuron $a_j$; $\overline{w}_{ij} \in \overline{W}$ where $\overline{W}$ is obtained by negating the model weights at the classification layer only. $\overline{P}^t(a_i|a_j)$ is only needed for \textit{contrastive} attention. 

We compute the neuron winning probabilities starting from the prior distribution encoding a given action/caption as follows:
\vspace{-2mm}
\begin{equation}
P^t(a_i) =\sum_{a_j\in \mathcal{P}_i}P^t(a_i|a_j)P^t(a_j) 
\label{eqn: main}
\end{equation}
\begin{equation}
\overline{P}^t(a_i) = \sum_{a_j\in \mathcal{P}_i}\overline{P}^t(a_i|a_j)\overline{P}^t(a_j) 
\label{eqn: main}
\end{equation}
\noindent where $\mathcal{P}_i$ is the set of parent neurons of $a_i$. 

\textbf{Temporal Normalization.} Replacing $tanh$ non-linearities with \textit{ReLU} non-linearities to extend EB in time does not suffice for temporal saliency. EB performs normalization at every layer to maintain a probability distribution. Hence, for spatiotemporal localization, signals from the desired $n^{th}$ time-step of a $T$-frame clip should be normalized in both time and space (assuming $S$ neurons in current layer) before being further backpropagated into the CNN: 
\vspace{-1mm}
\begin{equation}
P_N^t(a_i) = P^t(a_i) / \textstyle\sum\nolimits_{t=1}^{T} \textstyle\sum\nolimits_{i=1}^{S} P^t(a_i).
\label{eqn: main}
\end{equation}
\begin{equation}
\overline{P}_N^t(a_i) =  \overline{P}^t(a_i) / \textstyle\sum\nolimits_{t=1}^{T} \textstyle\sum\nolimits_{i=1}^{S} \overline{P}^t(a_i).
\label{eqn: main}
\end{equation}
$c$EB-R computes the difference between the normalized saliency maps obtained by EB-R starting from $O$, and EB-R starting from $\overline{O}$ using negated weights of the classification layer. $c$EB-R is more discriminative as it grounds the evidence that is unique to a selected class/word. For example, $c$EB-R of \textit{Surfing} will give evidence that is unique to \textit{Surfing} and not common to other classes used at training time (see Fig.~\ref{fig:pointgame} for an example). This is conducted as follows:
\vspace{-2mm}
\begin{equation}
\begin{split}
Map^t(a_i) = P_N^t(a_i) - \overline{P}_N^t(a_i).
\end{split}
\label{eqn: main}
\end{equation}
\indent \textbf{CNN Backward.} For every video frame $f_t$ at time step $t$, we use the backprop of \cite{zhang2016top} for all CNN layers: 
\begin{equation}
P^t(a_i|a_j) = 
\begin{cases}
    Z_j \widehat{a}_i^t w_{ij},     & \text{if } w_{ij}\geq 0,     \\
    0,                            & \text{otherwise}
\end{cases}
\label{eqn:CNN}
\end{equation}
\begin{equation}
Map^t(a_i) = \sum_{a_j\in P_i} P^t(a_i|a_j)Map^t(a_j) 
\label{eqn:final}
\end{equation}
%
\noindent where $\widehat{a}^t_i$ is the activation when frame $f_t$ is passed through the CNN. $Map^t$ at the desired CNN layer is the $c$EB-R saliency map for $f_t$. Computationally, the complexity of $c$EB-R is on the order of a \textit{single} backward pass. Note that for EB-R, $P_N^t(a_j)$ is used instead of $Map^t(a_j)$ in Eqn.~\ref{eqn:final}. The general framework applied to both video action recognition and captioning is summarized in Algorithm~\ref{alg1}. Details of each task are discussed in the following two sections.

\vspace{-1mm}
\subsection{Grounding: Video Action Recognition}
\label{grounding_vid_axn_recog}
\vspace{-1mm}

In this task, we ground the evidence of a specific action using a model trained on action recognition. The task takes as input a video sequence and the action ($\mathcal{A}$) to be localized, and outputs spatiotemporal saliency maps for this action in the video. We use the CNN-LSTM implementation of \cite{donahue2015long} with VGG-16 \cite{simonyan2014very} for our action grounding in video. This encodes the temporal information intrinsically present in the actions we want to localize. The CNN is truncated at the \textit{fc7} layer such that the \textit{fc7} features of frames feed into the recurrent unit. We use a single LSTM layer.



\begin{figure}[t]
\centering
\includegraphics[width=\linewidth, trim={0.3cm 0.4cm 0.8cm 0.6cm},clip]{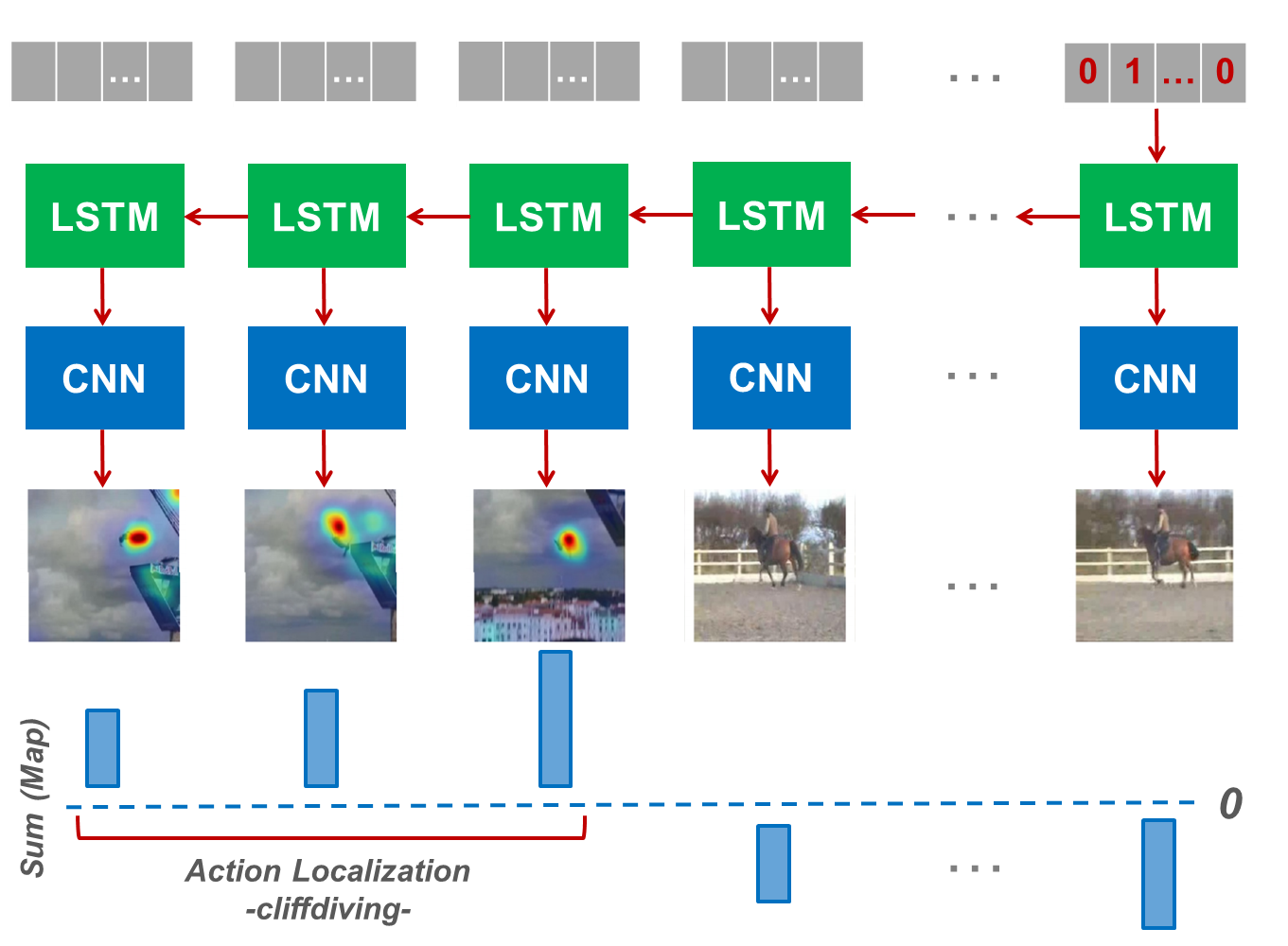}
\caption{\small{Grounding Action Recognition. The red arrows depict $c$EB-R for spatiotemporal grounding of the action \textit{CliffDiving}. 
Starting from the last LSTM time-step, $c$EB-R backpropagates the probability distribution through time and through the CNN at every time-step. The saliency map for each time-step is used for the spatial localization. The sum of each saliency map, over time, is then used for temporal localization of the action within the video, as described in Sec.~\ref{grounding_vid_axn_recog}.
}}
\label{fig:$c$EB-R}
\end{figure}

Performing $c$EB-R results in a sequence of saliency maps $Map^{t}$ for $t=1, ..., T$ at \textit{conv5} (various layers perform similarly \cite{zhang2016top}). These maps are then used to perform the temporal grounding for action $\mathcal{A}$. Localizing the action, entails the following sequence of steps. First, the sum of every saliency map is computed to give a vector $\mathcal{S} \in \mathbb{R}^{T}$. Second, we find an anchor map with the highest sum. Third, we extend a window around the anchor map in both directions in a greedy manner until a saliency map with a negative sum is found. A negative sum indicates that the map is less relevant to the action $\mathcal{A}$ under consideration. This allows us to determine the start and end points of the temporal grounding, $s_{\mathcal{A}}$ and $e_{\mathcal{A}}$ respectively.
Fig.~\ref{fig:$c$EB-R} depicts the $c$EB-R pipeline for the task of action grounding.

\vspace{-1mm}
\subsection{Grounding: Video Captioning} \label{captioning model}
\vspace{-1mm}

\begin{figure}[t!]
\centering
\includegraphics[width=\linewidth, trim={0.25cm 0.4cm 0.8cm 0.6cm},clip]{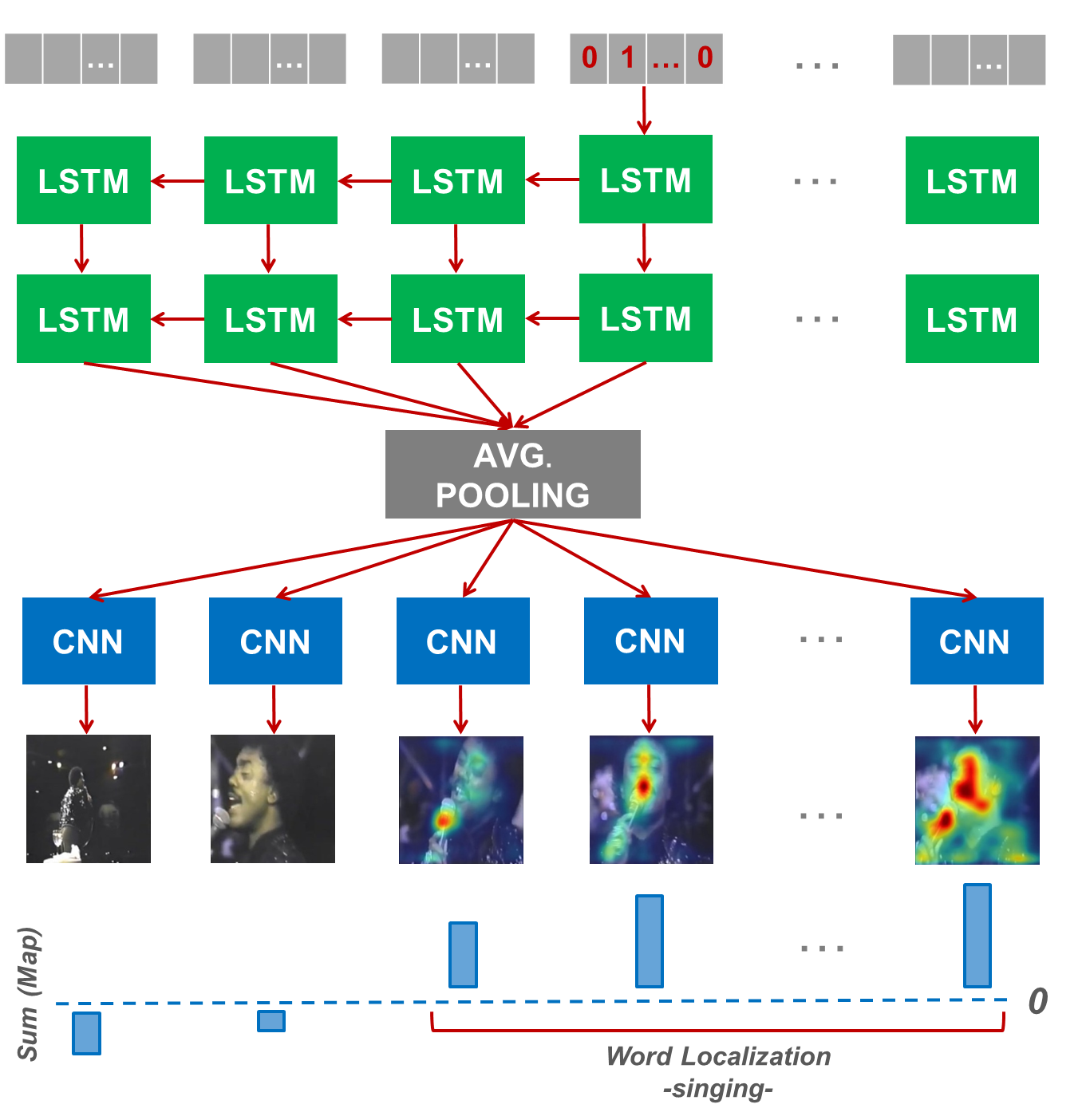}
\vspace{-1.7em}
\caption{\small{Grounding Captioning. The red arrows depict $c$EB-R for spatiotemporal caption grounding. The video caption produced by the model is \textit{A man is singing on a stage}. Starting from the time-step corresponding to the word \textit{singing}, $c$EB-R backprops the probability distribution through the previous time-steps and through the CNN. The saliency map for each time step is used for spatial localization.
The sum of each saliency map, over time, is then used for temporal localization of the word within the clip.}}
\label{fig:$c$EB-R_caption}
\end{figure}

In this task, we ground evidence of word(s) using a model trained on video captioning. The task takes as input a video and word(s) to be localized, and outputs spatiotemporal saliency maps corresponding to the query word(s). We use the captioning model of \cite{venugopalan2014translating} to test our $c$EB-R approach. This model consists of a VGG-16, followed by a mean pooling of the  VGG \textit{fc7} features, followed by a two-layer LSTM. Fig.~\ref{fig:$c$EB-R_caption} depicts $c$EB-R for caption grounding. 

We backpropagate an indicator vector for the words to be visualized starting at the time-steps they were predicted, through time, to the average pooling layer. We then distribute and backpropagate probabilities among frames -according to their forward activations (Eqn.~\ref{eqn:CNN})- through the VGG until the \textit{conv5} layer where we obtain the corresponding saliency map. Performing $c$EB-R results in a sequence of saliency maps $Map^{t}$ for $t=1, ..., T$ grounding the words in the video frames. Temporal localization is performed using the steps described in Sec.~\ref{grounding_vid_axn_recog}. 

\begin{figure}[H]
 \removelatexerror
 \begin{algorithm}[H]
	\caption{$c$EB-R}\label{alg1}    
	\vspace{.2 cm}
	\KwIn{$T$-frame video clip, pre-trained CNN-LSTM model, $\mathcal{A}$: action or word to be localized in the video.}
    \vspace{.2 cm}
	\KwOut{Spatial saliency maps of $\mathcal{A}: Map^{t}$ for $t = 1, ..., T$.}
    \vspace{.2 cm} 		 
	{\bf Procedure}:\vspace{.2 cm}\\
	\nl Set a one-hot vector according to the desired action class or caption word $\mathcal{A}$ at the desired $n^{th}$ time-step\;\vspace{.2 cm}
	\nl Backprop the indicator vector through time and down to the \textit{fc} CNN layer using EB-R obtaining a saliency map $Map^{t}$ at every time step $t$\;\vspace{.2 cm}
	\nl Normalize the resulting frame-wise saliency maps over time such that $\sum_{t=1}^{T}Map^{t} = 1 $\;\vspace{.2 cm}
	\nl Repeat the above steps, with negated weights at the top layer to get a second set of $T$ saliency maps\;	\vspace{.2 cm}
    \nl Contrastive Operation: Subtract the resulting maps at the \textit{fc} CNN layer to yield $c$EB for each time step\;\vspace{.2 cm}
    \nl Continue EB through the CNN to the desired \textit{conv} layer to obtain the spatial grounding\;\vspace{.2 cm}
     \nl The sum of each spatial saliency map over time can be used to perform temporal grounding for $\mathcal{A}$\;
     \label{alg:$c$EB-R}
 \end{algorithm}
\end{figure}


\section{Experiments: Action Grounding}
In this work we ground the decisions made by our deep models. In order to evaluate this grounding, we compare it with methods that localize actions. Although our framework is able to jointly localize actions in space and time, we report results for spatial localization and temporal localization separately due to the lack of an action dataset that has untrimmed videos with spatiotemporal bounding boxes. 


\subsection{Spatial Localization} \label{spatial localization}
\label{spatial_loc}

In this section we evaluate how well we ground actions in space. We do this by comparing our grounding results with ground-truth bounding boxes localizing actions per-frame. 

\textbf{Dataset.} \textit{THUMOS14} \cite{THUMOS14} provides per-frame bounding box annotations of humans performing actions for 
3207 videos of 24 classes from the \textit{UCF101} dataset \cite{soomro2012ucf101}. \textit{UCF101} is a trimmed video dataset containing 13320 actions belonging to 101 action classes. 

\textbf{Baselines.}  We compare our formulation against spatial top-down saliency using a CNN (treating every video frame as an independent image). We also compare against standard backpropagation (BP), and BP for RNNs (BP-R).

\textbf{Models.} We use the following CNN model: VGG-16 of Ma \etal \cite{ma2017less} trained on \textit{UCF101} video frames and BU101 web images for action recognition with a test accuracy of 83.5\%. We use the following CNN-LSTM model: the same VGG-16 fine-tuned with a one-layer LSTM on \textit{UCF101} for action recognition with a test accuracy of 83.3\%.


\textbf{Setup and Results.} We use the bounding box annotations to evaluate our spatial grounding using the pointing game introduced by Zhang \etal \cite{zhang2016top}. We locate the point having maximum value on each top-down saliency map.
Following \cite{zhang2016top}, if a 15-pixel diameter circle around the located point intersects the ground-truth bounding-box of the action category for a frame, we record a hit, otherwise we record a miss. We measure the spatial action localization accuracy by $Acc =\#Hits / (\#Hits+\#Misses) $ over all the annotated frames for each action. 

Table \ref{table:cnn_lstm_comparison} reports the results of the spatial pointing game. Extending top-down saliency in time (-R) consistently improves the accuracy for all three methods, compared to performing top-down saliency separately on every frame of the video using a CNN. EB-R has the greatest absolute improvement of 5.7\%. 

\begin{table}[t]
\centering
\begin{tabular}{MM|MM|MM} 
\hline
\multicolumn{6}{c}{\textbf{Method Acc (\%)}} \\
\hline
EB & EB-R & $c$EB & $c$EB-R & BP & BP-R \\ 
\hline
55.8 & \textbf{61.5} & 37.0 & \textbf{39.1} & 37.3 & \textbf{39.2} \\
\hline
\end{tabular}
\vspace{-0.5em}
\caption{\small{Accuracy of the spatial pointing game conducted on $\sim$3K videos of \textit{UCF101} for spatially locating humans performing actions in videos. The results show that extending top-down saliency in time (-R) improves the accuracy compared to performing top-down saliency separately on every frame of the video using a CNN. The non-contrastive versions work better for reasons described in the text.}}
\vspace{-1em}
\label{table:cnn_lstm_comparison}
\end{table}

\begin{figure}[t]
\centering
\begin{subfigure}[b]{\linewidth}
\centering
\includegraphics[width=0.49\textwidth, height=0.25\textwidth]{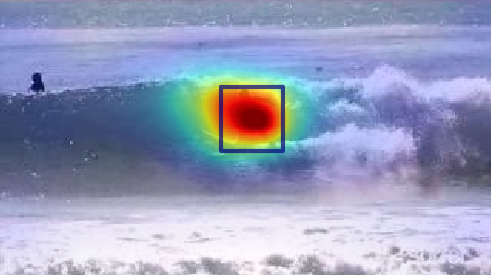}
\includegraphics[width=0.49\textwidth, height=0.25\textwidth]{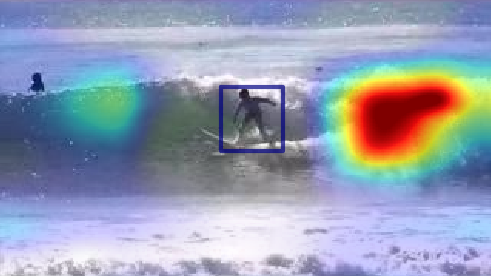}
\caption{\small{Grounding \textit{Surfing} using EB-R (L) and $c$EB-R (R)}}
\end{subfigure} 
\begin{subfigure}[b]{\linewidth}
\centering
\includegraphics[width=0.49\textwidth, height=0.25\textwidth]{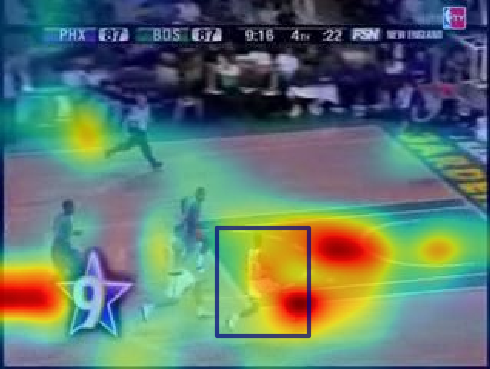}
\includegraphics[width=0.49\textwidth, height=0.25\textwidth]{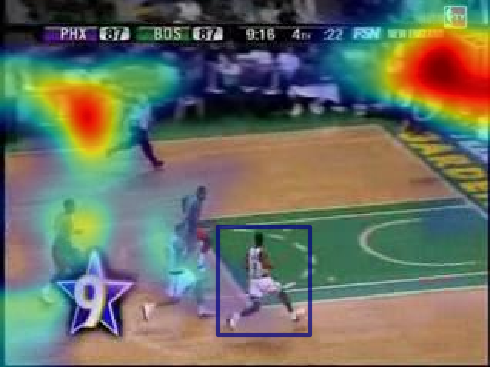}
\caption{\small{Grounding \textit{BasketballDunk} using EB-R (L) and $c$EB-R (R)}}
\end{subfigure}
\vspace{-1.7em}
\caption{\small{The saliency maps produced by EB-R (left) and $c$EB-R (right) together with the \textit{THUMOS14} groundtruth bounding box over the same frame of the actions (a) \textit{Surfing} and (b) \textit{BasketballDunk}. In both cases, EB-R highlights the most salient regions of the frame for this action (human), which matches the bounding box annotation. However, $c$EB-R highlights the region that is unique to the ground truth action: the waves for \textit{Surfing}, and the hoop for \textit{BasketballDunk}. This is because highlighting the human region does not provide insightful information to the classifier.}}
\vspace{-1em}
\label{fig:pointgame}
\end{figure}


We note that the non-contrastive versions outperform their contrastive counterparts. This is because they highlight discriminative evidence for actions, which may not necessarily be the humans performing the actions. For example, for many actions in \textit{UCF101}, the human may be in a standing position, in which case $c$EB-R will highlight cues that are discriminative and unique to this action rather than highlighting the human. These cues may belong to the context in which the activity is performed, or the action classes on which the model was trained. We demonstrate this in Fig.~\ref{fig:pointgame} for the actions \textit{Surfing} and \textit{BasketballDunk}.

\subsection{Temporal Localization}

In this section we evaluate how well we ground actions in time. We do this by comparing our grounding results with ground-truth action boundaries. 

\textbf{Datasets.} We first use a simple and controlled setting to validate our method by creating a synthetic action detection dataset. We then present results on the \textit{THUMOS14} \cite{THUMOS14} action detection dataset. 
The synthetic dataset is created by concatenating two \textit{UCF101} videos uniformly sampled: a ground truth (\textit{GT}) video, and a random (\textit{rand}) background video, such that class(\textit{GT}) $\neq$ class(\textit{rand}). The two actions are concatenated, first sequentially (\textit{rand} + \textit{GT} or \textit{GT} + \textit{rand}) in 16-frame clips, then inserted at a random position (\textit{rand} + \textit{GT} + \textit{rand}) in 128-frame clips. We use all 3783 test videos provided in \textit{UCF101}, each in combination with a different random background video. The \textit{THUMOS14} dataset consists of 1010 untrimmed validation videos and 1574 untrimmed test videos of 20 action classes. Among test videos, we evaluate our framework on the 213 test videos which contain annotations as in \cite{xu2017r,shou2017cdc}.

\textbf{Baselines.} For the synthetic experiment, we compare $c$EB-R and EB-R with a probability-based approach where we threshold the predicted probability (to $1$ if $\ge 0.5$, to $-1$ if $< 0.5$) of the \textit{GT} class at every time-step. For the detection experiment in \textit{THUMOS14} we compare our proposed method with state-of-the-art approaches. 

\textbf{Models.} For the synthetic dataset, we use the same CNN-LSTM model used for spatial action grounding (Sec.~\ref{spatial_loc}). For the \textit{THUMOS14} dataset we use a CNN-LSTM model: the same VGG-16 model used for spatial action grounding (Sec.~\ref{spatial_loc}) fine-tuned with a one-layer LSTM on \textit{UCF101} and trimmed sequences from \textit{THUMOS14} background and validation sets.

\begin{figure}[t]
\centering
\begin{subfigure}[b]{\linewidth}
\centering
\includegraphics[width=\textwidth]{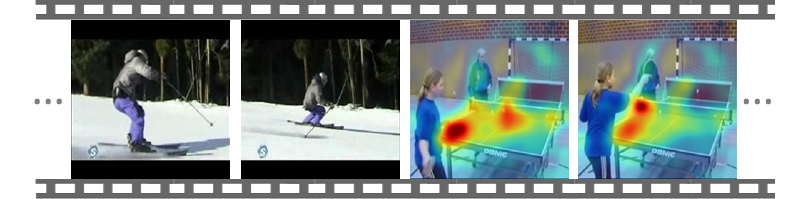}
\caption{\small{Grounding of the action \textit{TableTennisShot} in the video}}
\end{subfigure}
\begin{subfigure}[b]{\linewidth}
\includegraphics[width=\textwidth]{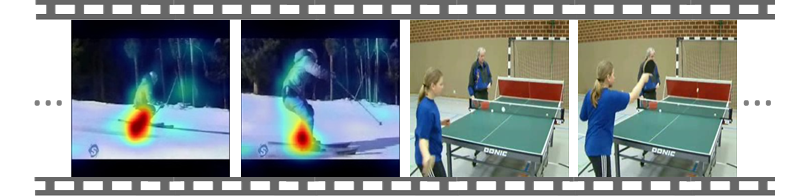}
\caption{\small{Grounding of the action \textit{Skiing} in the video}}
\end{subfigure}
\vspace{-1.7em}
\caption{\small{Applying \textit{contrastive} Excitation Backprop for Recurrent Networks ($c$EB-R) to produce spatiotemporal localization of actions in sample frames of a video. Demonstrated here is (a) $c$EB-R spatiotemporal localization of \textit{TableTennisShot} in a video (b) $c$EB-R spatiotemporal localization of \textit{Skiing} in the same video. The video consists of two consecutive actions that are synthetically concatenated: \textit{Skiing} followed by \textit{TableTennisShot}.}}
\vspace{-1em}
\label{fig:temporal_visualization}
\end{figure}

\begin{figure}[t]
\centering
\includegraphics[height=0.58\linewidth, width=\linewidth, trim={0.12cm 0.1cm 1cm 0.55cm},clip]{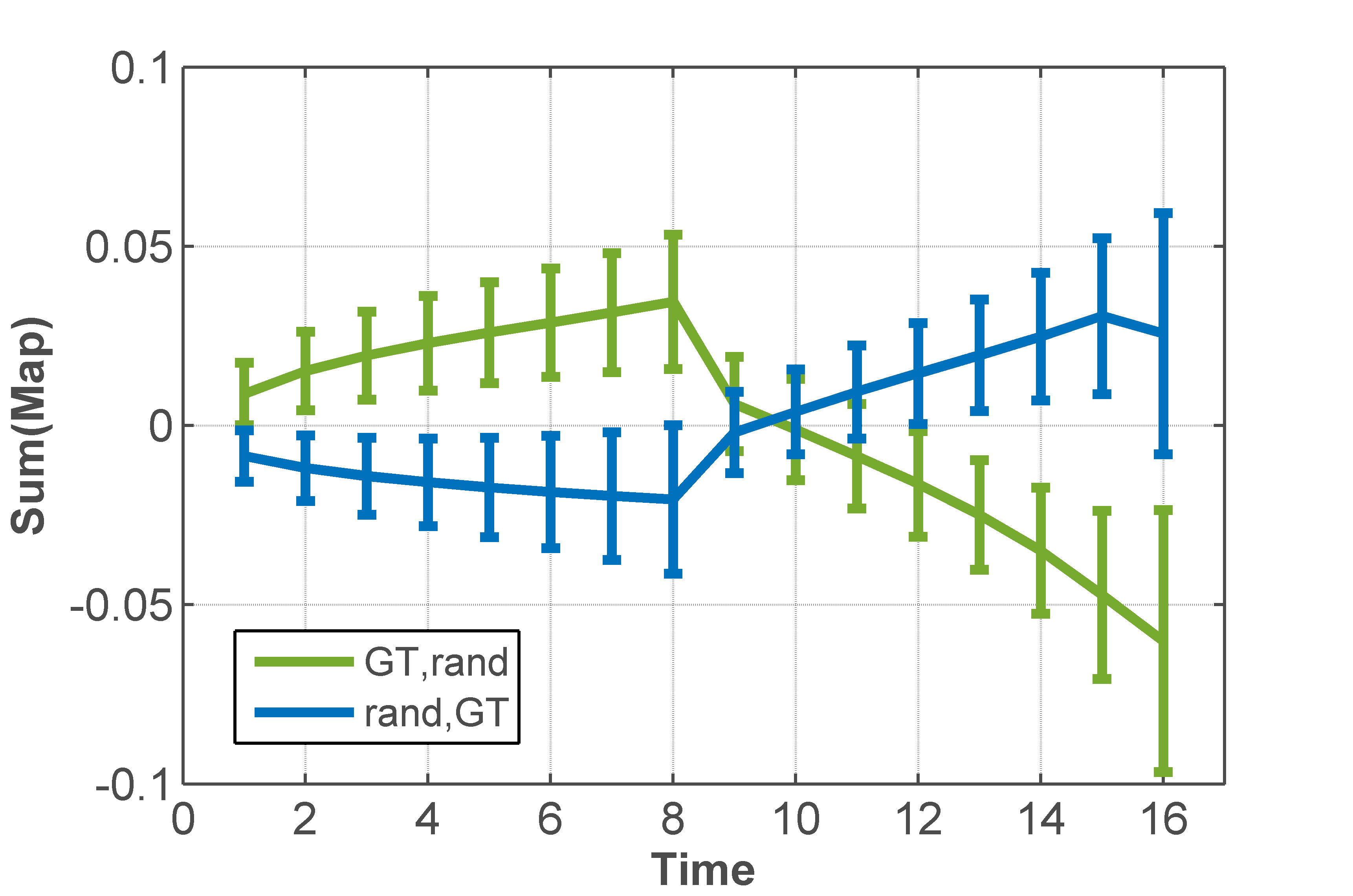}
\vspace{-1.7em}
\caption{\small{Sum of the saliency maps at \textit{fc7} over time, in frames, for synthetic videos that (blue) have a \textit{rand} action followed by a \textit{GT} action and (green) have a \textit{GT} followed by a \textit{rand} action. The average and standard deviation are reported over all test videos. $c$EB-R provides an accurate midway boundary between actions.}}
\label{fig:randbg_and_gt}
\end{figure}

\textbf{Setup and Results: Synthetic Data.}  
First, we perform experiments on the synthetic videos composed of two sequential actions, where the boundary is the midpoint. Fig.~\ref{fig:temporal_visualization} presents a sample spatiotemporal localization. The heatmaps produced by $c$EB-R correctly ground the queried action. More examples are presented in the supplementary material. While Fig.~\ref{fig:temporal_visualization} presents a qualitative sample, Fig.~\ref{fig:randbg_and_gt} quantitatively presents results on the entire test set. The action switches from \textit{GT} to \textit{rand} or vice versa midway. It can be seen that the sum of saliency maps is: positive and increasing as more of the \textit{GT} action is observed, negative and decreasing as more of the \textit{rand} action is observed.

Next, we perform experiments where we vary the length of the \textit{GT} action  that we want to localize inside a clip. To retain action dynamics, we sample \textit{GT} and \textit{rand} from the entire length of their corresponding videos. Table \ref{table:synthetic} presents the temporal localization results of our synthetic data. In the experimental setup with fixed action length we assume that we know the label and length of the action to be localized. 
To localize, we find the highest consecutive sum of attention maps for the desired action length. Regarding the sequences with unknown action lengths, we only assume the label of the action to be localized and perform the pipeline described in Sec.~\ref{grounding_vid_axn_recog}. In the bottom half of Table \ref{table:synthetic} we only report thresholded probabilities and $c$EB-R results since our localization procedure assumes negative values at action boundaries, whereas EB-R is non-negative. 
The grounded evidence obtained by $c$EB-R attains the highest detection scores, $73.5\%$ and $62.0\%$, for action sequences of known and unknown lengths, respectively, for IoU overlap between detections and ground-truth of $\alpha = 0.5$, despite the fact that the model is not trained for localization. 

\textbf{Setup and Results: \textit{THUMOS14} Pointing Game.} We evaluate the pointing game in time for \textit{THUMOS14} -a fair evaluation for methods that do not optimize for detection. For processing, we divide a video into 128-frame consecutive clips. We perform the pointing game by pointing \cite{zhang2016top} in time to the peak sum of saliency maps. For each ground-truth annotation we check if the detected peak is within its boundaries. If yes, we count it as a hit, otherwise, as a miss. We compare this approach with the peak position of predicted probabilities, and a random point in that clip. 

The results of this experiment are presented in Table \ref{table:POINTING GAME TIME}. Pointing to a random position clearly obtains lowest results while peak probability ($65.8 \%$) and $c$EB-R ($65.1 \%$) have similar performance. However, peak probability does not offer spatial localization. Peak probability uses the model prediction, while $c$EB-R uses the evidence of that prediction. Moreover, we observe that peak probability and $c$EB-R are complementary, yielding $77.4 \%$.

\textbf{Setup and Results: \textit{THUMOS14} Action Detection.}
We evaluate how well our grounding does on the more challenging task of action detection that it was not trained for. In this experiment, we divide a video into 128-frame consecutive clips for processing. Table \ref{table:real_data} presents the temporal detection results of the \textit{THUMOS14} dataset. Differently from the pointing game experiment, we detect the start and end of the ground-truth action. We note that although our method is not supervised for the detection task, we achieve an accuracy of 57.9\% when locating a ground truth class with an overlap $\alpha=0.1$ as demonstrated in Table \ref{table:real_data}. 

\section{Experiments: Caption Grounding}

In this section, we show how $c$EB-R is also applicable in the context of caption grounding. As observed by \cite{Ramanishka2017cvpr}, there is an absence of datasets with spatiotemporal annotations of frames for captions. Therefore, they propose the following experimental setup which we follow: qualitative results for the spatiotemporal grounding on videos, and quantitative results for spatial grounding on images. 

\textbf{Datasets.} We use the \textit{MSR-VTT} \cite{xu2016msr} dataset for video captioning and \textit{Flickr30kEntities} \cite{plummer2015flickr30k} for image captioning. 

\textbf{Models.} We use the CNN-LSTM-LSTM video captioning model of \cite{venugopalan2014translating} trained on \textit{MSR-VTT} to test our $c$EB-R approach for spatiotemporal grounding as described in Sec. \ref{captioning model}. We use the same video captioning model, without the average pooling layer, trained on \textit{Flickr30kEntities} for image captioning. The models have comparable METEOR scores to the Caption-Guided Saliency work of \cite{Ramanishka2017cvpr}, to which we compare our results: 26.5 (\vs 25.9) for video captioning and 18.0 (\vs 18.3) for image captioning.

\begin{figure}[t]
\centering
\begin{subfigure}[b]{\linewidth}
\centering
\includegraphics[width=\textwidth]{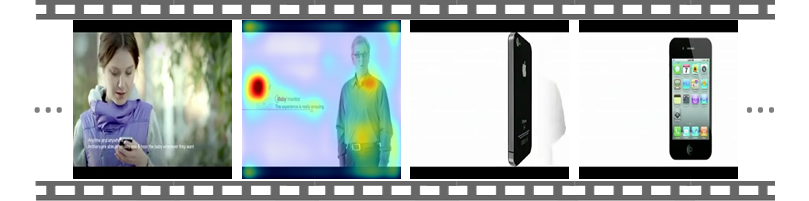}
\caption{\small{grounding of the word \textit{man}}}
\end{subfigure}
\begin{subfigure}[b]{\linewidth}
\includegraphics[width=\textwidth]{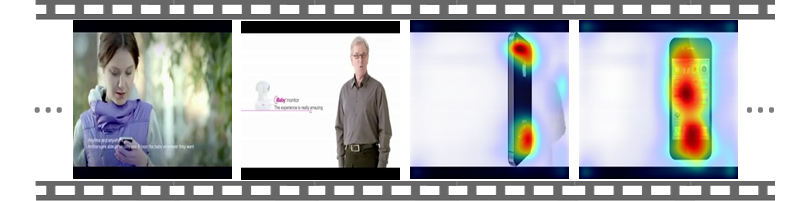}
\caption{\small{grounding of the word \textit{phone}}}
\end{subfigure}
\vspace{-1.7em}
\caption{\small{Comparison of grounding of words \textit{man} and \textit{phone} in the caption \textit{A man is talking about a phone} of a video from \textit{MSR-VTT} using $c$EB-R. The man is well localized in (a) and the phone is well localized in (b), as desired.}}
\label{fig:captioning_words}
\end{figure}

\textbf{Setup and Results.} 
For the \textit{MSR-VTT} video dataset, we sample 26 frames per video following \cite{Ramanishka2017cvpr} and perform grounding of nouns. Fig.~\ref{fig:captioning_words} presents the grounding for the word \textit{man} and \textit{phone} in the same video. The \textit{man} is well localized only in frames where a man appears, and the \textit{phone} is well localized in frames where a phone appears.

\begin{table}[h]
\centering
\begin{tabular}{cccccc} 
\hline
&&\textbf{\makecell{Length \\ (frames)}} & \textbf{\makecell{Probability \\ (\%)}}   & \textbf{\makecell{EB-R \\ (\%)}}  & \textbf{\makecell{$c$EB-R \\ (\%)}} \\ 
\hline \\[-0.6em]
{\multirow{6}{*}{\rotatebox[origin=c]{90}{\textbf{\textit{Action Length}}\hspace*{0.5em}}}} & {\multirow{3}{*}{\rotatebox[origin=c]{90}{\textbf{\textit{Known}}\hspace*{-0.5em}}}} & 11  &  8.5    &  11.3  &  \textbf{15.5} \\
&&41  &  28.2   &  38.5  &  \textbf{53.2} \\
&&65  &  47.7   &  56.3  &  \textbf{73.5} \\
\cdashline{2-6} \\[-0.5em]
&{\multirow{3}{*}{\rotatebox[origin=c]{90}{\textbf{\textit{Unknown}}\hspace*{-0.5em}}}}&11  & 3.4   &  -  & \textbf{ 4.1}  \\
&&41  &  9.5   &  -  &  \textbf{47.9}  \\
&&65  &  35.7   &  -  &  \textbf{62.0}  \\
\hline
\end{tabular}
\vspace{-0.5em}
\caption{\small{Action detection results on synthetic data, measured by mAP at IoU threshold $\alpha = 0.5$. Top part of table: methods assume that the length and label of the action to be detected are known. Bottom part of table: methods assume that the label is known, but the length is unknown. $c$EB-R attains best performance.
}}
\label{table:synthetic}
\end{table}

\begin{table}[h]
\centering
\begin{tabular}{l|c} 
\hline
\textbf{Method} & Accuracy (\%)  \\ 
\hline
Random & 57.3 \\
Peak probability & 65.8  \\
$c$EB-R &  65.1 \\
Peak probability + $c$EB-R & \textbf{77.4 } \\ 
         
\hline
\end{tabular}
\vspace{-0.5em}
\caption{\small{Pointing game in time performed on the \textit{THUMOS14} test set. The probability of an action together with the evidence for presence of the action are complementary and give a great improvement in accuracy when combined.}}
\vspace{-1em}
\label{table:POINTING GAME TIME}
\end{table}

\begin{table}[t]
\centering
\begin{tabular}{l|c} 
\hline
\textbf{\makecell{Method}} & \textbf{mAP ($\alpha = 0.1$)} \\ 
\hline
Karaman \etal \cite{karaman2014fast} & 4.6 \\
Wang \etal \cite{wang2014action} & 18.2 \\
Oneata \etal \cite{oneata2014lear} & 36.6 \\
Richard \etal \cite{richard2016temporal} & 39.7 \\
Shou \etal \cite{shou2016temporal} & 47.7 \\
Yeung \etal \cite{yeung2016end} & 48.9 \\
Yuan \etal \cite{yuan2016temporal} & 51.4 \\
Xu \etal \cite{xu2017r} & 54.5 \\
Zhao \etal \cite{SSN2017ICCV} & 60.3\\
Kaufman \etal \cite{Kaufman_2017_ICCV} & 61.1 \\
\hline
Ours 
&  57.9  \\
\hline
\end{tabular}
\vspace{-0.5em}
\caption{\small{Our approach \vs fully supervised approaches for action detection on \textit{THUMOS14}, measured by mAP at IoU threshold $\alpha = 0.1$. Although our model is not trained for action detection (trained for recognition), we achieve 57.9\%, which is comparable to state-of-the-art when localizing a ground truth action in a video.}}
\label{table:real_data}
\end{table}

\begin{table}[t]
\centering
\begin{tabular}{l|c} 
\hline
\textbf{\makecell{Method}} & \textbf{Avg (Noun Phrases)} \\ 
\hline
Baseline random & 0.268\\
Baseline center & 0.492\\
Caption-Guided Saliency \cite{Ramanishka2017cvpr} & 0.501\\
Ours &  \textbf{0.512} \\
\hline
\end{tabular}
\vspace{-0.5em}
\caption{\small{Evaluation of spatial saliency on \textit{Flickr30kEntities} using $c$EB-R. Baseline random samples the maximum point uniformly and Baseline center always picks the center.}}
\vspace{-1em}
\label{table:spatial_saliency}
\end{table}

We quantitatively evaluate our results of spatial grounding using the pointing game on the \textit{Flickr30kEntities} and compare our method to the Caption-Guided Saliency work of \cite{Ramanishka2017cvpr}, following their evaluation protocol. We use ground truth captions as an input to our model in order to reproduce the same captions. Then, we use bounding box annotations for each noun phrase in the ground truth captions and check whether the maximum point in a saliency map is inside the annotated bounding box. 



\begin{figure}[t]
\centering
\begin{subfigure}[b]{\linewidth}
\centering
\includegraphics[width=0.24\textwidth]{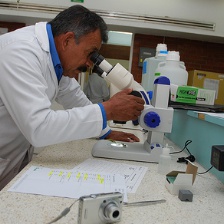}
\includegraphics[width=0.24\textwidth]{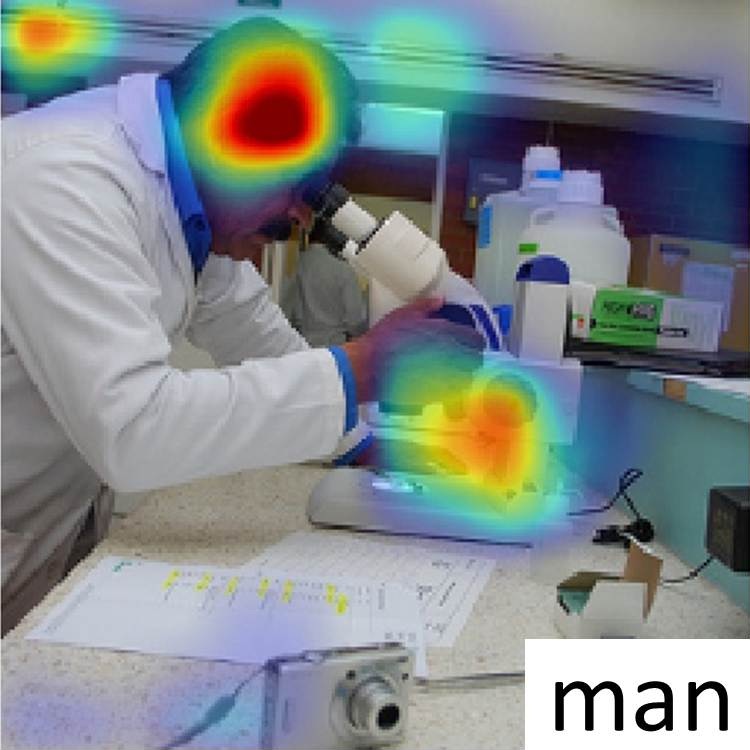}
\includegraphics[width=0.24\textwidth]{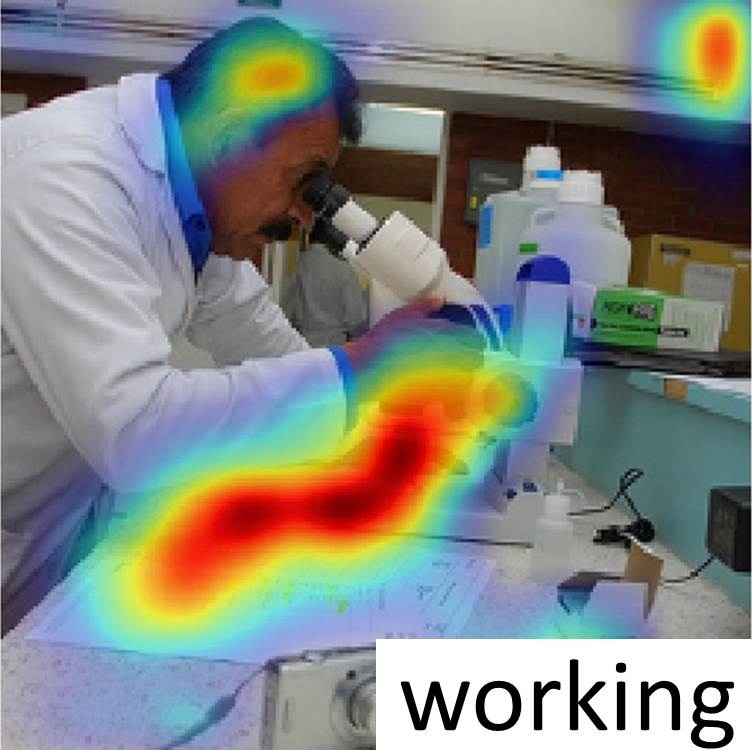}
\includegraphics[width=0.24\textwidth]{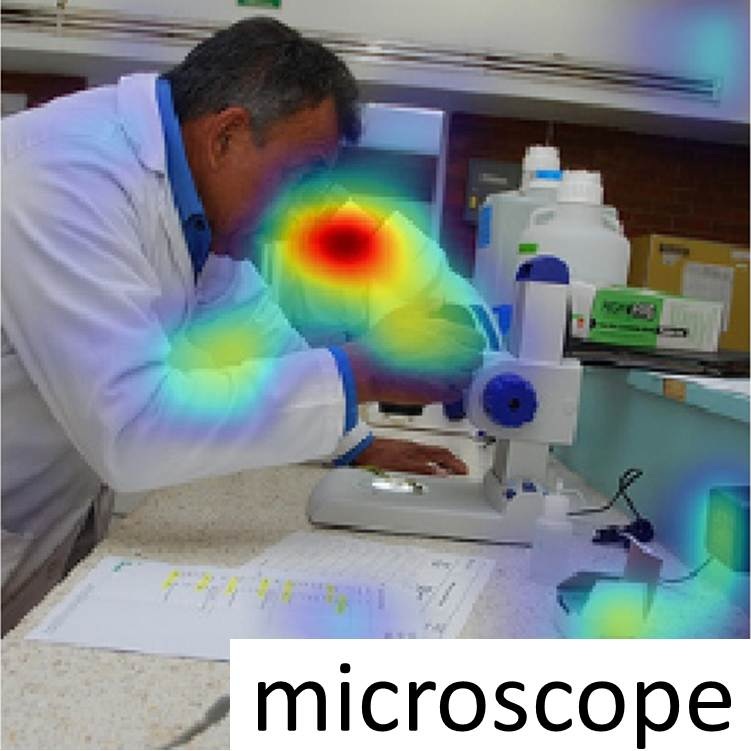}
\caption{\small{image caption: \textit{A man in a lab coat is working on a microscope.}}}
\end{subfigure}
\begin{subfigure}[b]{\linewidth}
\includegraphics[width=0.24\textwidth]{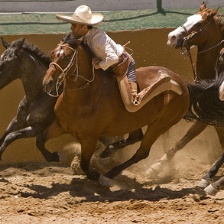}
\includegraphics[width=0.24\textwidth]{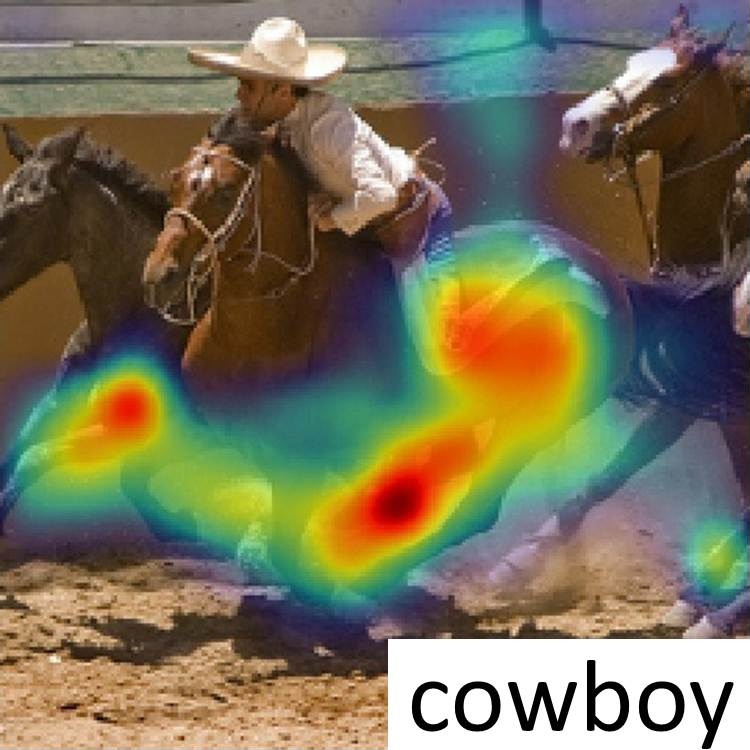}
\includegraphics[width=0.24\textwidth]{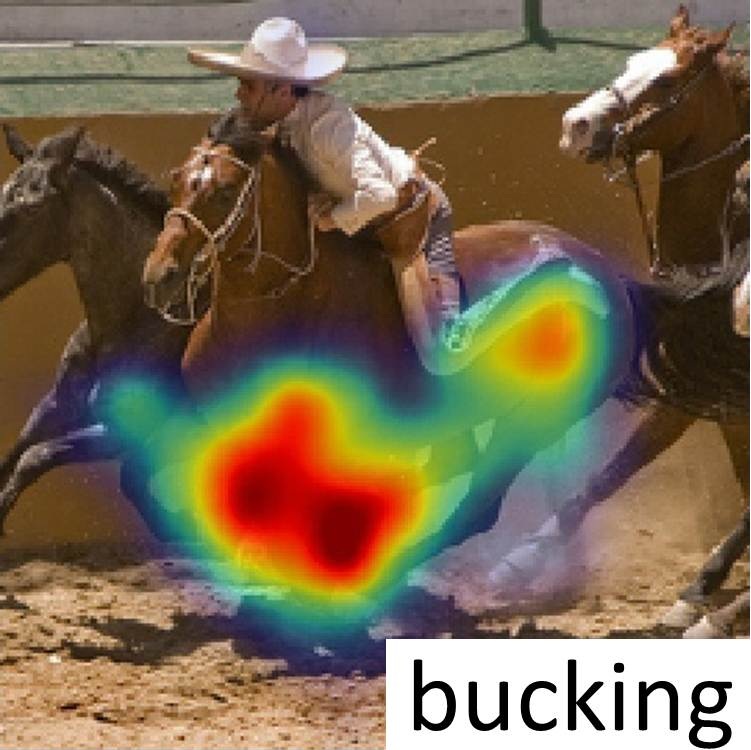}
\includegraphics[width=0.24\textwidth]{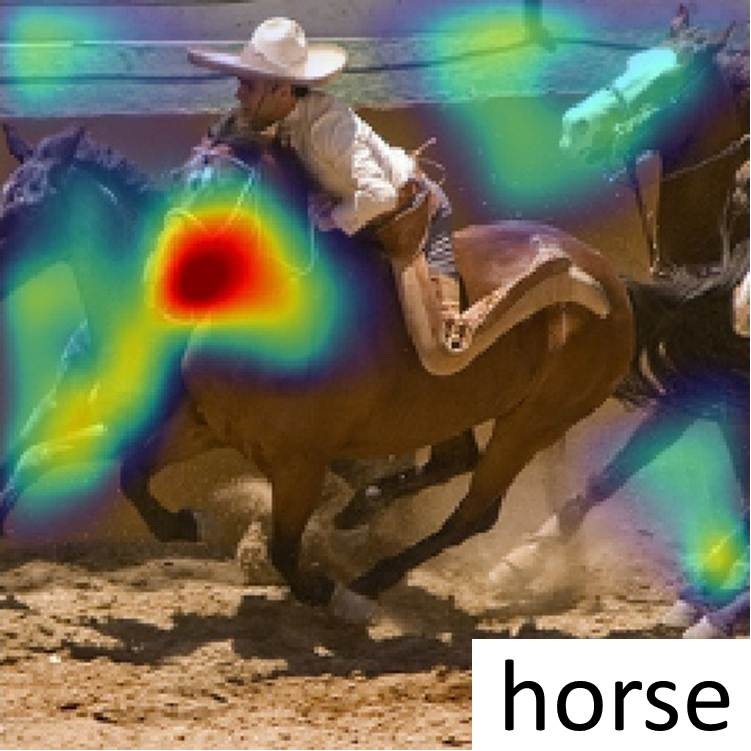}
\caption{\small{image caption: \textit{A cowboy is riding a bucking horse.}}}
\end{subfigure}
\vspace{-1.7em}
\caption{\small{Grounding different words of a caption using $c$EB-R for two images from the \textit{Flicker30kEntities} dataset.}}
\vspace{-0.5em}
\label{fig:flicker_qual}
\vspace{-2mm}
\end{figure}

Table \ref{table:spatial_saliency} shows the results of the spatial pointing game on \textit{Flickr30kEntities}. Our approach achieves comparable performance to \cite{Ramanishka2017cvpr}. In this experiment, we ground the ground truth captions to match the experimental setup in \cite{Ramanishka2017cvpr}. Although we follow their protocol for fair comparison, we note that our method can better highlight evidence using generated captions (\vs ground truth captions). This is because the evidence of a ground truth noun that is not predicted may not be sufficiently activated in the forward pass. Fig.~\ref{fig:flicker_qual} presents some visual examples of grounding in images using the generated captions.



 Our approach has a computational advantage over \cite{Ramanishka2017cvpr}. In order to obtain spatial saliency maps for a word in a video, $c$-EB-R requires one forward pass and one backward pass through the CNN-LSTM-LSTM, while \cite{Ramanishka2017cvpr} requires one forward pass through the CNN part, but $m$ forward passes through the LSTM-LSTM part, where $m=64$ is the area of the saliency map (\vs our single backward pass). Moreover, they require $f$ forward LSTM passes, where $f=26$ is the number of frames, to compute the temporal grounding, whereas ours is implicitly spatiotemporal.
 

\section{Conclusion}
In conclusion, we devise a temporal formulation, $c$EB-R, that enables us to visualize how recurrent networks ground their decisions in visual content. We apply this to two video understanding tasks: video action recognition, and video captioning. We demonstrate how spatiotemporal top-down saliency is capable of grounding evidence on several action and captioning datasets. These datasets provide annotations for detection and/or localization, to which we have compared the evidence in our generated saliency maps. We observe the strengths of $c$EB-R in highlighting discriminative evidence, which was particularly beneficial for temporal grounding. We also observe the strengths of its variant, EB-R, in highlighting salient evidence, which was particularly beneficial for spatial localization of action subjects. 

\vspace{-0.7mm}
\section*{Acknowledgments}
We thank Kate Saenko and Vasili Ramanishka for helpful discussions. This work was supported in part by NSF grants 1551572 and 1029430, an IBM PhD Fellowship, gifts from Adobe and NVidia, and Intelligence Advanced Research Projects Activity (IARPA) via Department of Interior/ Interior Business Center (DOI/IBC) contract number D17PC00341. The U.S. Government is authorized to reproduce and distribute reprints for Governmental purposes notwithstanding any copyright annotation thereon. Disclaimer: The views and conclusions contained herein are those of the authors and should not be interpreted as necessarily representing the official policies or endorsements, either expressed or implied, of IARPA, DOI/IBC, or the U.S. Government.

{\small
\bibliographystyle{ieee}
\bibliography{egbib}
}

\end{document}